\documentclass[10pt,conference]{IEEEtran}

\usepackage{tikz,pgfplots} 
\usepackage{booktabs, multicol, multirow}
\usepackage[acronym,nomain]{glossaries}
\usepackage[inline]{enumitem}
\usepackage{algorithm}
\usepackage{algorithmicx}
\usepackage{algpseudocode}
\usepackage{amsmath,amsfonts,amssymb}
\usepackage[font=small]{caption}
\usepackage{subfig}
\usepackage{grffile}
\usepackage{tabularx}
\usepackage{colortbl}
\usepackage{hhline}
\usepackage{hyperref}
\usepackage[flushleft]{threeparttable}
\usepackage{booktabs,fixltx2e}
\usepackage{balance}
\usepackage{boldline}
\usepackage{rotating}
\usepackage{pdflscape}
\usepackage[many]{tcolorbox}
\usepackage{framed}
\usepackage{cite}
\usepackage{flushend}
\usepackage{listings}
\usepackage{graphicx}
\usepackage{graphics}
\pgfplotsset{compat=1.14}

\usetikzlibrary{shapes.misc}

\tikzset{cross/.style={cross out, draw=black, minimum size=2*(#1-\pgflinewidth), inner sep=0pt, outer sep=0pt}, cross/.default={3pt}}

\colorlet{shadecolor}{blue!20}

\renewcommand{\mathbf}[1]{{\boldsymbol #1}}
\hypersetup{
	colorlinks   = false,
	citecolor    = blue,
	linkcolor    = blue,
}
\setlist[itemize]{leftmargin=*}
\setlist[enumerate]{leftmargin=*}

\pgfplotsset{ 
  legend style = {font=\small\sffamily}, 
  label style = {font=\small\sffamily} 
}

\usepackage[colorinlistoftodos,prependcaption,textsize=tiny]{todonotes}
\newcommand{\secref}[1]{Section~\ref{sec:#1}}

\newcommand{\figref}[1]{Figure~\ref{fig:#1}}

\DeclareCaptionFormat{nice}{\noindent\colorbox{blue!10}{\hspace{1em}#1#2#3\hspace{1em}}}

\makeglossaries

\begin{document}
%


\lstdefinestyle{interfaces}{
  float=tp,
  floatplacement=tbp}

\captionsetup[lstlisting]{format=nice,singlelinecheck=false,margin=3pt,
     font={bf,sf,footnotesize},skip=-5pt}


\title{Machine Learning Meets Quantitative Planning: Enabling Self-Adaptation in Autonomous Robots}

\author{\IEEEauthorblockN{Pooyan Jamshidi}
\IEEEauthorblockA{
University of South Carolina, USA}
\and
\IEEEauthorblockN{Javier C\'{a}mara}
\IEEEauthorblockA{
University of York, UK}
\and
\IEEEauthorblockN{Bradley Schmerl, Christian K{\"a}stner, David Garlan}
\IEEEauthorblockA{
Carnegie Mellon University, USA}}

\maketitle
\begin{abstract}

Modern cyber-physical systems (e.g., robotics systems) are typically composed of physical and software components, the characteristics of which are likely to change over time.
Assumptions about parts of the system made at design time may not hold at run time, especially when a system is deployed for long periods (e.g., over decades). 
Self-adaptation is designed to find reconfigurations of systems to handle such run-time inconsistencies. Planners can be used to find and enact optimal reconfigurations in such an evolving context. However, for systems that are highly configurable, such planning becomes intractable due to the size of the adaptation space.
To overcome this challenge, in this paper we explore an approach that (a) uses machine learning to find Pareto-optimal configurations without needing to explore every configuration and (b) restricts the search space to such configurations to make planning tractable.
We explore this in the context of robot missions that need to consider task timeliness and energy consumption. 
An independent evaluation shows that our approach results in high-quality adaptation plans in uncertain and adversarial environments. 




\end{abstract}


\section{Introduction}
\label{sec:introduction}

Modern software-intensive systems often incorporate components that are likely to change their behavior
over time (\emph{e.g.}, third-party web services whose performance or availability varies due to changes in implementation or software controllers in cyber-physical systems in which reliability progressively decrease due to wear and tear of hardware). 
In such systems, software might not be able to continue operation as expected when the assumptions made at design time about constituent parts of the system do not hold at run time. 
This issue also affects self-adaptive systems, in which the effects of adaptations might become progressively degraded with respect to their {expected behavior}. 

Robotics software is a particularly good candidate in studying how self-adaptive mechanisms can be compromised when assumptions about constituent parts of the system break, because these assumptions are oftentimes more brittle than in other domains~\cite{kramer2007self}. Examples include those about software components associated with hardware elements like sensors that degrade over time, becoming less accurate or consuming more energy than expected. 

To characterize expected system behavior, existing approaches in self-adaptive systems almost always rely on manual modeling by domain experts~\cite{DBLP:journals/scp/CamaraLGS16}, which is expensive and potentially unreliable, or they use small and artificial models that need to be adjusted as the system evolves~\cite{JVKSK:SEAMS17}. 
Regardless of the approach employed, the vast space of configurations and environmental conditions of such systems almost invariably lead to strong simplifying assumptions and models that do not accurately reflect important interactions among components, configuration options, and environmental variables.  

\begin{figure}[t!]
\centering
\includegraphics[width=\columnwidth]{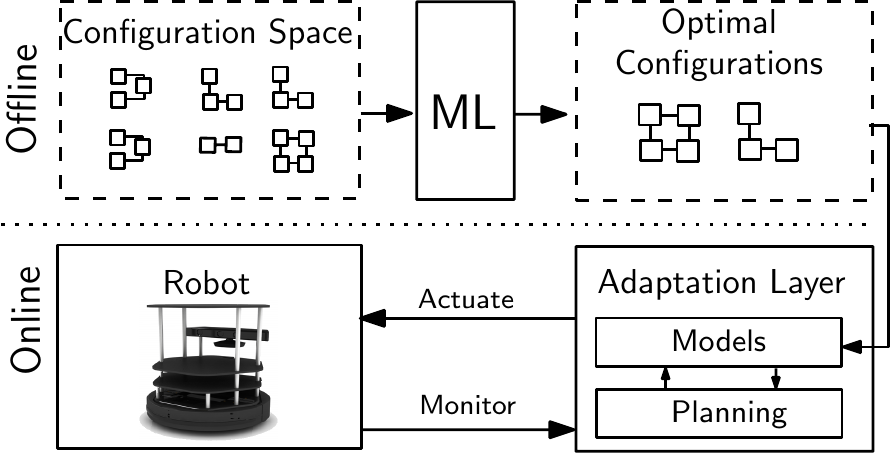}
\caption{Overview of our approach: We use machine learning to identify optimal configurations that will be used for run-time adaptations.}
\label{fig:overview}
\end{figure}

To make a concrete case, let us consider an example of a model that informs the robot about its power consumption under different configurations (cf. \figref{overview}). This model can help to determine a new configuration that the robot should adapt to in order to cope with environmental uncertainties. If the robot detects, for example, that it is running low on energy, it may choose to use a less accurate but more efficient algorithm to determine its location in order to preserve battery life for the rest of the mission. However, the effectiveness of this adaptation will only be as good as the underlying assumptions on which decision-making relies such as accurate estimation of energy consumption. If, for example, the energy demand of the configuration is higher than expected, the robot may run out of battery and fail to complete its mission.   


In this work, we consider robotics software as a highly-configurable system, in which system characteristics (\emph{e.g.}, usage of sensors) are treated as configuration options~\cite{ABC:Software,JVKSK:SEAMS17}. Our approach can, therefore, choose from many configurations of the robot at run time based on environmental situations. For instance, a robot may choose to decrease the localization accuracy to preserve the battery when it passes through an area where less sensory input is needed.

To choose an appropriate configuration, a self-adaptive system must come up with a plan. Although 
planners can be used to decide the best way to adapt the configuration at run time~\cite{DBLP:conf/sigsoft/MorenoCGS15,DBLP:conf/aips/Lacerda0H17}, a key challenge to overcome when tackling highly-configurable systems is that the configuration space is exponentially large in the number of configuration options of the system, which can consist of millions or even billions of possible configurations~\cite{JVKSK:SEAMS17}. In such situations, planning, based on exhaustive search, becomes intractable due to the size of the adaptation space. To overcome this challenge, 
the core idea behind our approach is to find interesting configurations that provide optimal performance and also constrain planning to only those configurations. We argue that this can make the planning problem tractable for run-time adaptation. We use machine learning to find Pareto-optimal configurations of the system with respect to the mission goals (\emph{e.g.}, energy consumption and timeliness) and pass only those configurations to the planner to constrain its search space. The novelties of our approach are as follows: (i)~the integration of learning and quantitative planning to enable run-time self-adaptation of highly-configurable systems, and (ii)~the integration of information from multiple heterogeneous models in quantitative planning, which enables reasoning that accounts for multiple quality aspects (e.f., energy consumption and timeliness) and their interactions. 
In this paper, we explore the following hypotheses:

\begin{shaded*}

\noindent \textbf{Hypothesis 1}: Machine learning can find optimal configurations without needing to explore all configurations.


\smallskip
\noindent \textbf{Hypothesis 2}: Restricting online planning to search only Pareto-optimal configurations leads to tractable run time planning that still results in high-quality adaptation plans.



\end{shaded*}

These hypotheses are not obvious; instead, it is possible that the space of configurations are too large and performance behavior is too complex that learning with a limited number of measurements cannot find optimal configurations and that the result of using a learned approximation will not yield adequate plans. This paper tests these hypotheses in the context of a highly-configurable robot performing missions in an interior environment. We use established machine learning techniques to identify Pareto-optimal configurations offline from observations of a sampling of system executions (\figref{overview}). We then exploit the product of learning by using it as one of the inputs to a planner that is used to reason quantitatively about the outcome of adaptation decisions in the wider context of the robot's mission and its objectives, including tradeoffs among qualities like timeliness and energy efficiency.


Our results show that our approach allows automated run-time decision-making for self-adaptation of highly-configurable systems by enabling a deeper exploration of what otherwise are intractable solution spaces. Specifically, the main contributions of this paper are as follows: 
\begin{itemize}
\item A \emph{learning} approach that reduces the planning space for systems that have many possible configurations in a multi-dimensional trade-space. 
\item An \emph{integration} of the learning with online model-based quantitative verification to enable timely planning for self-adaptive systems.
\item An \emph{open source implementation} of the infrastructure that allows us to evaluate the approach in a robotics adaptation scenario. This infrastructure enabled independent third party evaluation (conducted by MIT Lincoln Laboratory\footnote{\url{https://www.ll.mit.edu/}}), and we discuss the results of this as well as lessons learned.
\item Evaluations via \emph{controlled experiments} for demonstrating the effectiveness of the learning approach as well as the scalability of the online planning.
\end{itemize}

\section{Running Example}
\label{sec:example}

We illustrate the details of our approach on a robotics adaptation scenario where the robotic system is considered as a \emph{highly-configurable} system and the adaptation concerns \emph{changing the configuration in which the robot is operating at run time}. Mobile robotic systems---sophisticated combinations of physical hardware and control software that sense and move through an environment---are increasingly relied upon in modern society to provide support in domains like healthcare, transportation and logistics, environmental protection, and assessment and maintenance of infrastructures.
In these domains, mobile robotic systems are expected to perform a wide variety of long range, long term, and complex missions. In contrast to industrial robotics, where a continuous energy supply can be assumed in most scenarios, the management of autonomous mobile robot missions requires being able to predict the energy consumption of the robot accurately to avoid undesirable situations like running out of energy  mid-way through a mission. Moreover, the combined space of the characteristics of the mission, the conditions of the operating environment, and the different robot configurations (sensors, actuators, computation intensive control algorithms, etc.) exhibits a high degree of variability, making energy consumption difficult to predict and manage. 

\begin{figure}
      \centering
      \includegraphics[width=0.8\linewidth]{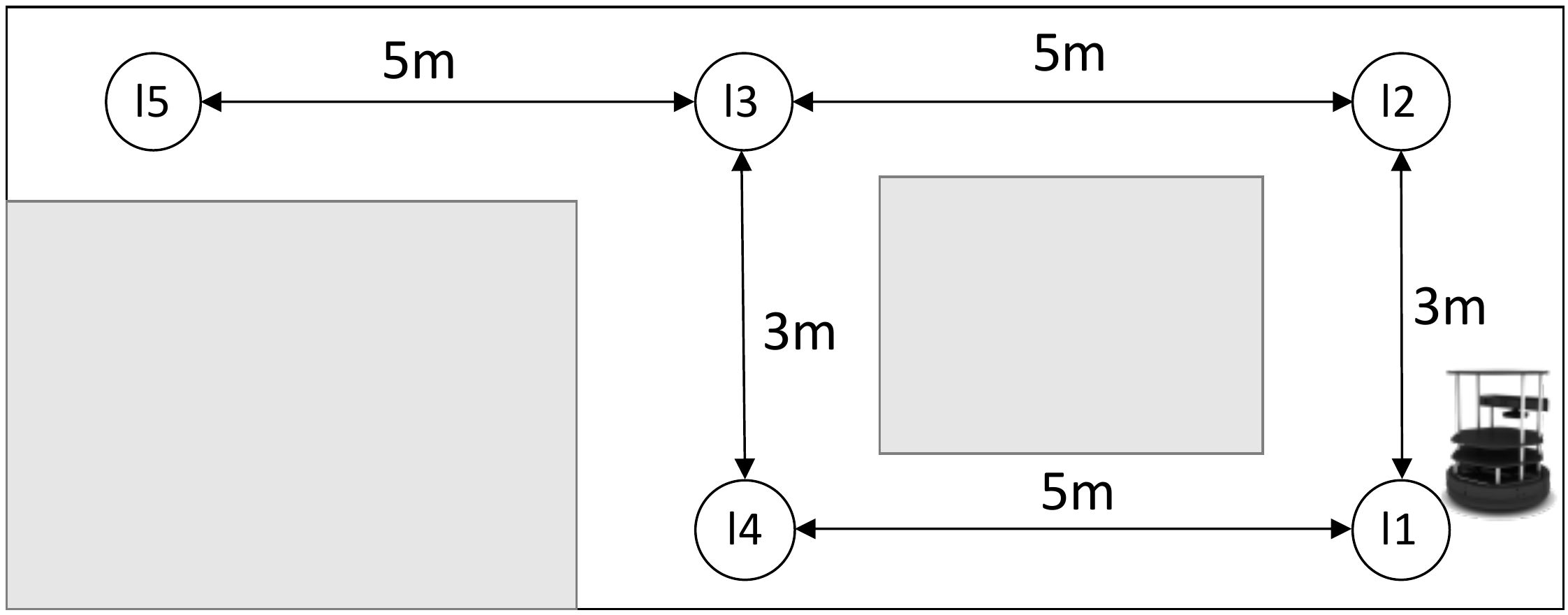}
   \caption{Simple mobile robotics scenario where a robot need to accomplish multiple tasks by traversing an uncertain terrain.}\label{fig:scenario}
\end{figure}

Figure~\ref{fig:scenario} shows a Turtlebot,\footnote{\url{https://www.turtlebot.com/}} which runs ROS (Robotic Operating System)\footnote{ROS (\url{http://www.ros.org}) is a framework for writing robot software that consists of a set of tools, libraries, and conventions that aim to simplify the task of creating complex and robust robot behavior across robotic platforms.} compatible components for different robotics tasks, \emph{e.g.}, localization components that determine the position of the robot in a map. These components are highly-configurable~\cite{ABC:Software}; for example, in Turtlebots we can enable/disable a sensor (\emph{e.g.}, Kinect) or change the number of particles used by the localization\footnote{\url{http://wiki.ros.org/amcl}}~\cite{KK:WSR16}. 
In this scenario, the mission of the robot is, starting at some location, to navigate to several locations in sequence to pick up and deliver materials in the shortest possible time with a limited battery.
During the execution of the mission, the robot has to adapt to some unpredictable environmental events that include (i)~encountering obstacles that make a particular segment impassable and (ii)~sudden changes in the battery's energy level (simulating sensor, battery degradation, or terrains that require more energy to navigate). To adapt in the face of these changes in order to avoid failing the mission, the robot can reconfigure (\emph{e.g.}, to a less energy-demanding configuration) and change the path (perhaps to reach a charging station).\footnote{See \url{https://www.youtube.com/watch?v=ec6BhQp2T0Q}}


\section{Overview of the Approach}
\label{sec:intuition}

The overall goal of our approach is to enable self-adaptation in highly-configurable systems that operate in dynamic and uncertain environments, using configuration change as the main mechanism to enact adaptation. The key challenge in this context is to overcome the size of the configuration space by finding a configuration of the system that yields a suitable adaptation for a situation given at run time.
Our intuition is that not all configurations of a highly-configurable system are important. In other words, there are dominant configurations with respect to a system's characteristics to which we can constrain the exploration (we call this \emph{the interesting part} of the configuration space). Therefore, instead of wasting resources on searching for an adaptation plan in the whole space of configurations, we can achieve good results if we perform exhaustive search only on optimal configurations. 

Figure~\ref{fig:overview} illustrates the overall structure of our approach, which is divided into two phases: {\bf (i)~{\em offline}}
, in which we use machine learning to learn a performance model~\cite{JVKSK:SEAMS17,JVKS:FSE18} of the system and identify optimal configurations and {\bf (ii)~{\em online}}, in which we use planning based on quantitative verification~\cite{DBLP:journals/cacm/CalinescuGKM12} to decide the best adaptation based on the set of optimal configurations identified in the design time phase.

\begin{figure}[t]
	\begin{center}
		\includegraphics[width=0.8\columnwidth]{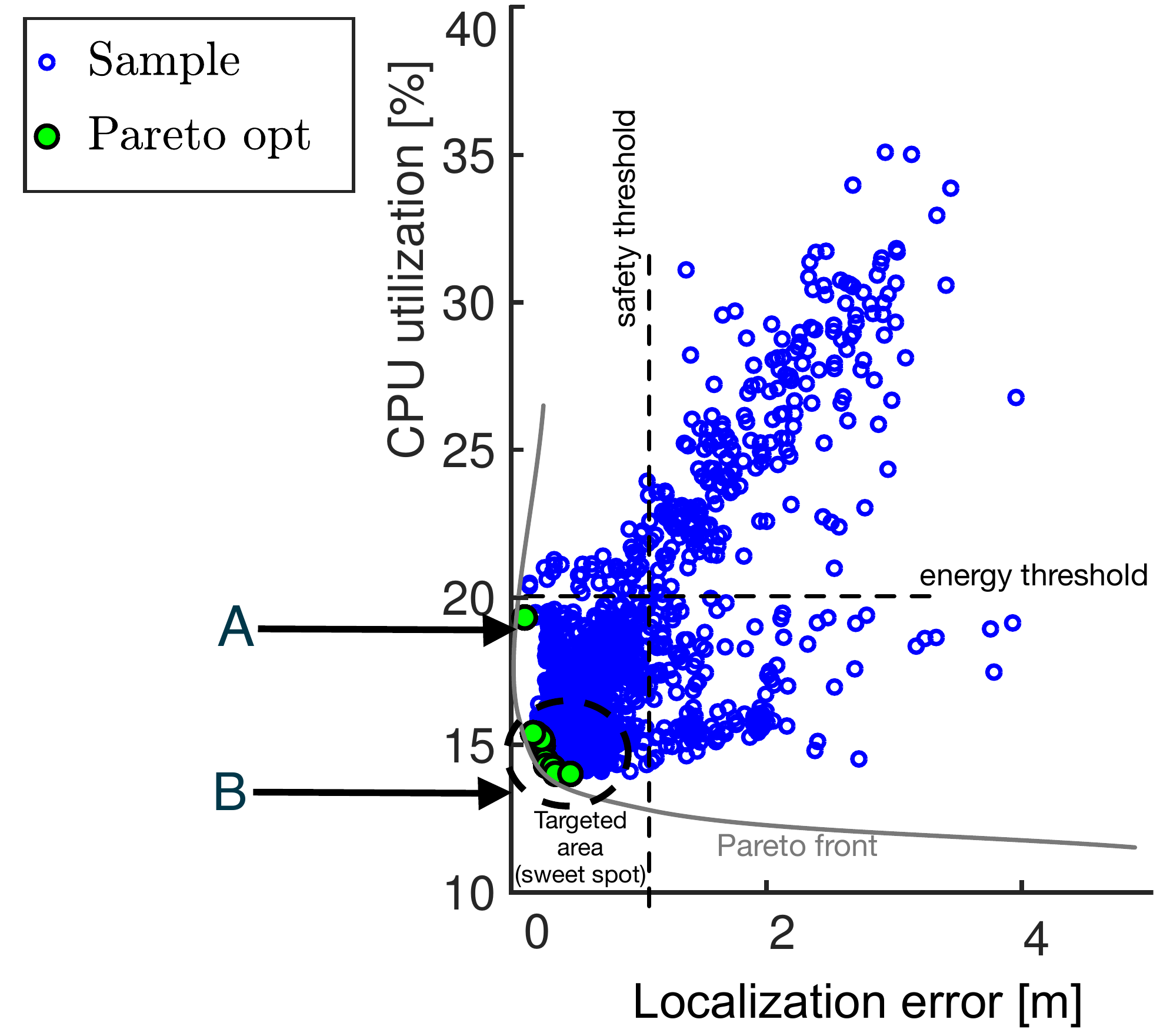}
		\caption{Pruning of adaptation space by selecting Pareto front configurations of the robot.}
		\label{fig:motivation}
	\end{center}
\end{figure}

The key insight of our approach is that we reduce the adaptation search space, in the offline phase, by incorporating information about the optimal configurations into the models that the adaptation layer employs for planning in the online phase.
To illustrate how our approach works in practice, Figure~\ref{fig:motivation} shows a few thousand configurations of our robotic system in which the performance of each configuration is measured in terms of multiple objectives including localization error (distance) and the percentage of CPU utilization (a proxy for energy consumption).
Offline, our approach can identify the Pareto-optimal configurations (highlighted) using learning. 
At run time, if adaptation is triggered, because, for example, the robot suddenly detects that it is running low on energy when it is in configuration {\sf A}, the planner can find an adaptation that includes an alternative (\emph{e.g.}, less energy-demanding) configuration by looking only among the Pareto-optimal (green) ones in region {\sf B} (which is significantly constrained subset compared to the overall space of configurations represented by the blue dots in the figure).  
This reduction in the size of the adaptation space leads to drastically reducing the planning time without compromising the quality of adaptations. 

\section{Integrated model discovery and self-adaptation}
\label{sec:rapproach}

This section provides the core technical contribution of this paper. We start by describing the learning mechanism (\secref{learning}) that facilitates finding Pareto-optimal configurations. We then proceed with adaptation planning (\secref{adaptation}) that enables quantitative reasoning about the outcome of adaptation decisions. Finally, we discuss the infrastructure (\secref{integration}) that integrates all components together and enables end-to-end adaptation. 


\subsection{Model Learning}
\label{sec:learning}


In this section, we introduce our learning approach to finding Pareto-optimal configurations that are used in planning. We first define some \emph{terms} to formalize the problem of \emph{model learning} followed by detailing the \emph{learning machinery} as well as the way we selected \emph{Pareto-optimal configurations}.

\subsubsection{Definitions}

Let $O_i$ indicate the $i$-th configuration option, which ranges over a finite domain $Dom(O_i)$. In general, $O_i$ may either indicate (i) an integer variable (\emph{e.g.}, the \emph{number of iterative refinements} in a localization algorithm), (ii) a categorical variable (\emph{e.g.}, local vs global \emph{localization method}), or (iii) binary options (\emph{e.g.}, enabling/disabling \emph{depth sensors}). The \emph{configuration space} is a Cartesian product of the domains of the parameters of interest $\mathbb{C}=Dom(O_1) \times \dots \times Dom(O_d)$, where $d$ is the total number of configuration options (or dimension of the space). A configuration $\mathbf{c}$ then is a vector in the configuration space $\mathbf{c} \in \mathbb{C}$ that assigns a particular value along each dimension. For instance, $\mathbf{c}=<1,1,0,0,0>$ represents a configuration with 5 binary options, where the former two options are enabled and the latter three are disabled.


To predict performance of the system under different configurations, (i) we first build a model using measurements and learning, and 
then (ii) we use the learned model to find Pareto-front configurations with respect to its objectives.

\subsubsection{Model learning}

We formulate model learning as a multi-objective optimization problem in a configuration space $\mathbb{C}$ given constraints. Formally, we introduce a set of inequality constraints $c(\mathbf{c}) = (c_1(\mathbf{c})<=b_1,..., c_n(\mathbf{c})<=b_n), b = (b_1,\cdots,b_n)$ to the multi-objective optimization. For example, if a robot is configured to use a light-sensitive sensor when traversing a part of the environment with many windows, this may cause high sensor errors and lead to large localization errors, meaning that the robot cannot successfully localize itself and so cannot find a path to the target.
We define $f:\mathbb{C}\rightarrow\mathbb{R}$ to be our vector of objective functions $f=(f_1,\cdots,f_o)$ (\emph{e.g.}, $f_1$: energy, $f_2$: speed of a robot). In practice we know about $f$ via measurements, \emph{i.e.},~$y_i=f(\mathbf{c}_i),\mathbf{c}_i \in \mathbb{C}$. Our goal is to identify the Pareto frontier of $f$; that is, the set $\Gamma \subseteq \mathcal{C}$ of configurations which are not dominated by any other configurations (cf., \figref{motivation}), \emph{i.e.}, the maximally desirable $\mathbf{c}$ which cannot be optimized further for any single objective without reducing the value of some other dimension. We aim to approximate $\Gamma$ with the fewest possible measurements, since there is an exponentially large number of configurations. 

Our main goal is to learn a reliable regression model, $\hat{f}(\cdot)$, that can predict the performance of the system, $f(\cdot)$, given a limited number of observations $\mathcal{D}=\{(\mathbf{c}_i,y_i)\}$. Specifically, we aim to minimize the \emph{prediction error}:
\begin{equation} \label{eq:objective}
\arg \min pe= \Sigma_{\mathbf{c}\in\mathbb{C}} |\hat{f}(\mathbf{c})-f(\mathbf{c})|
\end{equation}




To learn a model that represents the performance landscape of the system, we build a regression model using \emph{stepwise linear regression}~\cite{H:Stepwise} from observations $\mathcal{D}$. We use stepwise regression for two reasons: It captures influential options and interactions, in an iterative manner, that \emph{scales} to a high-dimensional space~\cite{R:Technometrics}; and it provides a model that end users can verify and can be reliably used for run-time decision making. If the measurement data is not representative enough to learn a credible model (by estimating the prediction power of the learned model), we could measure more configurations. 


A regression model is a polynomial whose terms determine the performance of the system. Each term may refer to one or more options ($o_i \in \mathcal{O}$), describing the influence of that option or an interaction ($o_io_j$)~\cite{SGAK:ESECFSE15}:
\begin{align}
\label{eq:influence-model}
f(o_1,\cdots,o_d) = \beta_0 + \sum_{o_i \in \mathcal{O}} \beta_i o_i + \sum_{o_i,o_j \in \mathcal{O}} \beta_{i..j}(o_i..o_j),
\end{align} 
where $\beta\in \mathbb{R}$ represents the coefficients of the model, $\beta_i o_i$ represents the performance impact of individual options, and $\beta_{i..j}(o_i..o_j)$ represents the performance impact for interactions among multiple options (comprising not only quadratic terms, but also higher order terms up to the number of individual options). In this work, we assume configuration options are binary, so if an option appears in the performance model, the option is influential. Note that numeric and categorical options can be transformed into binary options by selecting two extreme values corresponding to zero and one; though in this transformation, we sacrifice precision~\cite{JVKS:FSE18}.  
Since the appearance of a term in the model is based on a statistical analysis, the structure of the model gives us a direct means to identify influential options and interactions. 


\subsubsection{The mechanics of learning}

We use both \emph{forward selection} and \emph{backward elimination} to learn the model~\cite{H:Stepwise}. Specifically, we use the p-value of an F-statistic~\cite{FHT:Book} to decide whether to add a term to the model or remove one. Specifically, the learning includes the following steps:
\begin{enumerate}
\item \emph{Initialization}: A model is fit to the data, and then the explanatory power of the model will be compared incrementally by forward selection and backward elimination. 
\item \emph{Forward Selection}: If any terms (options, \emph{e.g.}, $o_1$, or their interactions, \emph{e.g.}, $o_1o_2$) not in the model have p-values less than an entrance threshold (we set it to 0.05), add the one with the smallest p-value to the model and repeat this step, otherwise proceed to the next step. 
\item \emph{Backward Elimination}: If any model terms have p-values larger than an exit threshold (we set it to 0.05), remove the one with the largest p-value and go to the previous step. 
\item \emph{Termination}: The extraction process stops when neither (2) nor (3) improve the model. 
\end{enumerate}


The final model includes only terms that are statistically significant with the level of our choice (based on our choice of threshold for p-value).
As an example, consider the robot in our running example has two configuration options: $o_1$ (enabling/disabling Kinect) and $o_2$ (enabling/disabling Localization). After the learning is finished, we have the following model: 
\begin{align}
\label{eq:influence-model-example}
f(\cdot) = 2 + 3o_1 + 20o_2 + 17o_1o_2
\end{align}
This model (constructed synthetically) shows that enabling Localization consumes much more energy than enabling Kinect, and both of these two options are interacting; meaning that the power consumption of the robot is larger than the consumptions of each of the Kinect and Localization components individually. Note for learning the model, we implemented a standard approach~\cite{H:Stepwise}, but the novelty lies in the integration with the quantitative planning that will be introduced later.


\subsubsection{Selecting Pareto front configurations}

In the model learning problem, the multiple objectives (\emph{i.e.}, energy, speed) are conflicting (\emph{e.g.}, configurations under which the robot moves quickly will consume more energy than ones moving slowly) so that finding a single configuration that is the optimum of all objectives is unlikely. Once we learn a model for both objectives, we select configurations that lie on the Pareto front of the objectives. A configuration $\mathbf{c}_1$ is said to dominate $\mathbf{c}_2$ if $\forall i \in \{ 1\cdots m \}, f_i(\mathbf{c}_1) <= f_i(\mathbf{c}_2) \& \exists j \in \{ 1\cdots m \}, f_j(\mathbf{c}_1) < f_j(\mathbf{c}_2)$. A configuration is Pareto-optimal if it is not dominated by any other configuration in the configuration space and dominates at least one point. The Pareto-optimal set in the configuration space is called the Pareto set, and the set of Pareto-optimal configurations in the objective space is called Pareto front. The procedure for identifying Pareto-optimal configurations includes enumerating all configurations and checking whether the selected configuration is dominated by any other configuration in the configuration space. If the answer is no, this configuration will be put into the Pareto set; otherwise, it will be discarded.




\definecolor{prismgreen}{rgb}{0, 0.6, 0}

\lstdefinelanguage{Prism}{ 
        basicstyle=\color{black}\scriptsize\sffamily, 
        keywords= {bool,C,ceil,const,ctmc,double,dtmc,endinit,endmodule,endrewards,endsystem,F,false,floor,formula,G,global,I,init,int,label,max,mdp,min,module,nondeterministic,P,Pmin,Pmax,prob,probabilistic,R,rate,rewards,Rmin,Rmax,S,stochastic,system,true,U,X},
        keywordstyle={\bfseries\color{black}},
        numberstyle=\tiny\color{black},
        belowcaptionskip=\baselineskip,
        comment=[l] {//}, morecomment=[s]{/*}{*/}, 
        commentstyle= \color{prismgreen}, 
        tabsize=4, 
        captionpos=b, 
        escapechar=@ 
}

\subsection{Planning and Adaptation}
\label{sec:adaptation}


Adaptation analysis and planning in the robotics domain poses a challenge in terms of integrating information from heterogeneous models (with different semantics and representations) that capture different facets of the domain (\emph{e.g.}, energy consumption, physical space, and safety).

We tackle this challenge by incorporating in our approach planning that abstracts relevant pieces of information from different models and integrates them into high-level models amenable to quantitative verification. 
We use the model checker PRISM in the planner's back-end to reason quantitatively about the outcome of adaptation decisions in a rich trade-off space.
Using this technology is particularly convenient for the following reasons: (a)~it is equipped with a high-level modeling language that helps in bridging the representation and semantic gap across multiple heterogeneous models, and (b)~it enables quantitative verification of properties that can be encoded as rewards/costs (\emph{e.g.}, energy, time).

This approach is distinguished from our other work reported in~\cite{DBLP:conf/sac/CamaraGS015,DBLP:journals/scp/CamaraLGS16,DBLP:conf/icse/CamaraMG15,DBLP:conf/sigsoft/MorenoCGS15,DBLP:journals/taas/CamaraMGS16,DBLP:journals/ase/CamaraSMG18}, in which we use quantitative probabilistic model checking for planning purposes, because it is the first one able to integrate information from heterogeneous models into richer formal specifications that allow capturing mutual dependencies among multiple aspects of the domain (\emph{e.g.}, energy consumption, speed, etc.). 

\smallskip
We illustrate our approach using a simple scenario (Figure~\ref{fig:scenario}), in which we assume that the mission of the robot is to navigate to a target location (${l5}$) from an initial location (${l1}$) in the shortest possible time with a limited battery.
To achieve this goal, the robot can move between locations and change its configuration.
Note that carrying out actions in different configurations might yield different results. For instance, moving between two locations at a higher speed is good in terms of timeliness but might consume more energy (these two aspects may also vary between configurations that have different sets of sensors enabled). This is an important aspect that enables the system to make trade-offs, like reconfiguring to reach its target location earlier if it has an energy surplus in the battery or choosing a more energy-efficient configuration while sacrificing timeliness if the battery is low on energy.

When synthesizing a specification for the robot behavior to complete the mission, the planner considers two main concerns of our scenario: (i)~{\em timeliness}---the robot should get to the target in the shortest possible time, and (ii)~{\em efficiency}---the robot should try to get to the target location with as much remaining energy in the battery as possible.

 These specifications can be used to produce an initial plan for the mission as well as for adapting at run time and synthesizing a new plan whenever there is a situation that demands it. Specifically, we consider two types of situations that demand adaptation in this scenario: (i)~the robot is unexpectedly running low on battery and might not have enough energy to finish the mission and (ii)~the robot is not able to make progress in going through a corridor (\emph{e.g.}, because there is an obstacle that cannot be circumvented).
 

To inform our synthesis process, we have four models that describe different aspects of the domain.

\begin{enumerate}

\item{\em Physical Environment}. Describes the physical space that the robot is navigating. We assume that this model is captured as a graph, where nodes correspond to physical locations, and arcs correspond to trajectories between them. Arcs are tagged with a {\em distance} attribute (5m and 3m for horizontal and vertical arcs in Figure~\ref{fig:scenario}, respectively).

\item{\em Architecture}. Captures the modes in which the robot can operate, as determined by the components that can be employed at run-time (\emph{e.g.}, different sensors, different navigation algorithms), their connections, and configuration parameters (\emph{e.g.}, speed, localization accuracy).

\item{\em Operations}. Includes the repertoire of behavior primitives that the robot can execute to carry out an action in the physical environment. For simplicity, we assume the robot can carry out three types of action: (i)~move between two locations, (ii)~change the robot's configuration, and (iii)~charge its battery when placed in a charging station.  

\item{\em Power}. This model is queried to determine: (i)~the available resources (\emph{i.e.}, the amount of energy remaining in the battery), and (ii)~the cost of operations in a given configuration (\emph{e.g.}, how much energy it takes to move to a given location at a given speed with a given set of sensors). 

\item {\em Task}. Describes the task to accomplish, and its progress. In this scenario, we assume that the progress of the mission captures the location of the robot, its remaining battery, the history of actions executed, and the remainder of actions to be executed.

\end{enumerate}

Analyzing properties of the system requires combining pieces of information captured in different models, which do not provide insight if considered individually. For example, the topology of the environment, the catalog of operations, or the power model are not useful on their own to obtain metrics for timeliness and energy efficiency when planning the robot's mission, but they can be combined to estimate the time and energy that the robot will take to complete its mission. To the best of our knowledge, our quantitative planning technique is the first that takes a systematic approach to provide such insights. The approach is divided in:

\subsubsection{Projection} Each problem domain model is projected into a view that abstracts its important features in an intermediate language. This translation is model type-specific--for each type of model a translator is written. For example, an architecture configuration view exposes a set {\em confs}, and a power view exposes a function $\mathsf{pow}(bp, c)$ that returns the power needed to carry out behavioral primitive $bp$ (\emph{e.g.}, move between locations {\small \sf $l1$} and {\small \sf $l2$}) in configuration $c \in$ {\em confs}. Thus, the power view can be related to the configuration view. 

\subsubsection{Planning model generation} The views are then combined by an aggregator that composes information from the different views (\emph{e.g.}, computing the energy consumed when a robot moves between two locations requires information from the environment, architecture, and power models). 

We carry out model integration into formal specifications based on the syntax of the PRISM language~\cite{KNP11} for Markov Decision Processes (MDP). 
Using MDP as the underlying formalism for our models allows underspecification of choices via nondeterminism, which can be resolved by the model checker via policy synthesis, enabling automated plan generation.
A PRISM MDP model is built as a set of processes or {\it modules} (delimited by keywords {\small \sf module/endmodule} that are encoded as a set of commands: 

\smallskip
\centerline{[$action$] $guard$ $\rightarrow$ $p_1:u_1$+ $\dots$ + $p_n:u_n$,}
\smallskip

where $guard$ is a predicate over the model variables (which can be either boolean or bounded-range integers, c.f., Listing~\ref{lst:robot}, lines 4-5). Each update $u_i$ describes a transition that the process can make (by executing $action$) if the guard is true. An update is specified by giving the new values of the variables, and has an assigned probability $p_i \in [0,1]$.\footnote{Probabilities enable convenient encoding of aleatoric uncertainties that the system might be subject to, \emph{e.g.}, of collision against obstacles in corridors. However, probabilistic aspects of modeling/reasoning are not core to the discussion in this paper, and hence are left out of scope for the sake of clarity. The interested reader may refer to other works for a comprehensive description of probabilistic planning in self-adaptive systems~\cite{DBLP:conf/sac/CamaraGS015,DBLP:conf/sigsoft/MorenoCGS15}.}

\smallskip 

Listing~\ref{lst:robot} shows the encoding for the robot module. The specification shows the state variables for the battery energy level ({\small \sf b}), the robot's location ({\small \sf l}), software configuration ({\small \sf c}), and heading ({\small \sf r}).
The commands that encode the operations to change the robot's configuration are lines 11-12, and the actions of the robot to move between locations are specified in~(lines 8-9).
Commands {\small \sf l1\_to\_l2} and {\small \sf l1\_to\_l4} (lines 11 and 12) have overlapping guards that introduce nondeterminism. This nondeterminism will be resolved later by PRISM by synthesizing a policy that specifies which actions should be chosen in different states to achieve the system's goal.

\begin{lstlisting}[style=interfaces,
			 language=Prism,
			 numbers=left,
			 frame=lines,
			 rulesepcolor=\color{black},
			 rulecolor=\color{black},
			 breaklines=true,
			 columns=fullflexible,
			 xleftmargin=0.5cm,
			 label=lst:robot,
			 caption={Robot module definition.},
			 belowcaptionskip=-7pt,
			 escapechar=$
			 ]
formula b_upd_l1_l2 = c=C_HALF_SPEED? max(0,b- $\colorbox{red}{357}$) : c=C_FULL_SPEED? max(0,b- $\colorbox{red}{628}$) : 0;
...
module bot_module
b:[0..MAX_BATTERY] init INITIAL_BATTERY; l:[0..5] init INITIAL_LOCATION;
c:[C_HALF_SPEED..C_FULL_SPEED] init C_HALF_SPEED;  r:[0..8] init INITIAL_HEADING; robot_done:bool init false;
	 [] true & (turn=RT) & (!stop) & (robot_done) -> (robot_done'=false) & (turn'=ET);
	 // Speed setting change tactics
	 [t_set_c_half_speed] (c!=C_HALF_SPEED) & (!stop) & (turn=RT) & (!robot_done) ->  (c'=C_HALF_SPEED) & (robot_done'=true);
	 [t_set_c_full_speed] (c!=C_FULL_SPEED) & (!stop) & (turn=RT) & (!robot_done) ->  (c'=C_FULL_SPEED) & (robot_done'=true);
	 // Robot navigation
	 [l1_to_l2] (l=l1) & (!stop) & (turn=RT) & (!robot_done) -> (l'=l2)  & (b'=b_upd_l1_l2) & (r'=H_SOUTH) & (robot_done'=true);
	 [l1_to_l4] (l=l1) & (!stop) & (turn=RT) & (!robot_done) -> (l'=l4)  & (b'=b_upd_l1_l4) & (r'=H_WEST) & (robot_done'=true);
  	 ...
	 [l5_to_l3] (l=l5) & (!stop) & (turn=RT) & (!robot_done) -> (l'=l3)  & (b'=b_upd_l5_l3) & (r'=H_EAST) & (robot_done'=true);
endmodule
\end{lstlisting}

The guard of {\small \sf l1\_to\_l2} is composed by conjunction of the following predicates: (i)~{\small \sf l=l1} specifies that the command can fire only if {\small \sf l1} is the current location of the system (represented by variable {\small \sf l});
(ii)~{\small \sf !stop} specifies that the command can fire only if the stop condition has not been achieved. The stop condition in this case is specified by the predicate {\small \sf (goal $|$ b $<=$ MIN\_BATTERY)}, where {\small \sf goal} encodes the fact that the robot has reached its target location ({\small \sf l = TARGET\_LOCATION}), and the second predicate captures the fact that the robot does not have enough battery to operate ({\small \sf b} represents the remaining battery energy);
(iii)~{\small \sf turn=RT} specifies that it is the robot's turn to act (variable {\small \sf turn} alternates turns between the system and the environment); and
(iv)~{\small \sf !robot\_done} specifies that the robot has not finished its turn to act yet.

When executed, each command updates state variables that include the robot's location ({\small \sf l}) and the battery's energy level~({\small \sf b}). {\em The battery level update that captures battery depletion is an important example of how different models have to be integrated to generate a useful formal specification of the system}. The battery update {\small \sf (b'=b\_upd\_l1\_l2)} refers to the formula defined in line 1 that encodes the depletion of the battery (357 milliwatt-hours) if the configuration is the one in which the setting is {\small \sf half} speed and a different (higher) one for {\small \sf full} speed. To compute these battery depletion values, translation has to integrate pieces of information from different models: (i)~from the environment model, the translator needs the distance $d_{l1-l2}$ between {\small \sf l1} and {\small \sf l2}; (ii)~it then computes the time it takes to move between those two locations as $t_{l1-l2} = d_{l1-l2} / s_{\sf c}$ (where $s_{\sf c}$ is the speed that corresponds to the current configuration ${\sf c}$); and (iii)~it then computes the battery depletion value by {\em piecing together the time needed to move between locations with information about the configurations supplied by learning (battery discharge rate)}, as $t_{l1-l2}*dr_{c}$, where $dr_{\sf c}$ is the battery discharge rate for configuration ${\sf c}$.

	\smallskip 
	
To calculate the overall qualities of the system, translation employs {\em reward/cost structures} that assign a reward/cost (a rational number) when commands are fired in the system. Listing~\ref{lst:timereward} shows a time reward/cost structure used to quantify the time required to complete the robot's mission. 
Line 4 specifies that, whenever command {\small \sf l1\_to\_l2} is fired, the {\small \sf time} {cost} accrues times that depend on the speed of the robot's configuration and the rotation time to make the robot head in the direction required to go from {\small \sf l1} to {\small \sf l2}. 

\begin{lstlisting}[style=interfaces,
			 language=Prism,
			 numbers=left,
			 frame=lines,
			 rulesepcolor=\color{black},
			 rulecolor=\color{black},
			 breaklines=true,
			 columns=fullflexible,
			 xleftmargin=0.5cm,
			 label=lst:timereward,
			 caption={Time reward definition.},
			 belowcaptionskip=-7pt,
			 escapechar=!
			 ]
formula rot_time_l1_to_l2 = r=H_NORTH ? 10.4720 : r=H_NORTHEAST ? 7.8540 : r=H_EAST ? 5.2360 : r=H_SOUTHEAST ? -2.6180 : r=H_SOUTH ? 0.0000 : r=H_SOUTHWEST ? 2.6180 : r=H_WEST ? 5.2360 : r=H_NORTHWEST ? 7.8540 :  0;
...
rewards "time"
	[l1_to_l2] true :c=C_HALF_SPEED? !\colorbox{red}{8.5714}! + rot_time_l1_to_l2 : !\colorbox{red}{4.4118}! + rot_time_l1_to_l2;
	...
	[l5_to_l3] true :c=C_HALF_SPEED? !\colorbox{red}{14.2857}! + rot_time_l5_to_l3 : !\colorbox{red}{7.3529}! + rot_time_l5_to_l3;
endrewards
\end{lstlisting}

\subsubsection{Synthesis} The learning model obtained as described in \secref{learning} provides a set of configurations in the Pareto 
front, including their energy discharge rate and speed, which are used to generate the parts of the task planning specification highlighted in red in Listings~\ref{lst:robot} and~\ref{lst:timereward}, respectively.
Hence, the planner generates a task plan specific to the set of Pareto-optimal configurations (the range of timeliness and energy efficiency levels vary with the configurations available at different points in the task's execution). 
Task plan generation is achieved via MDP {\em policy synthesis}. Synthesizing and checking for the existence of a {\em policy} (also called {\em strategy} or {\em adversary}) that is able to optimize an objective expressed as a quantitative property in probabilistic computation-tree logic (PCTL) extended with rewards/costs~\cite{DBLP:conf/formats/AndovaHK03} is a fundamental aspect of model checking MDPs. An objective expressed in a PCTL property can state that an MDP has a policy that can ensure that an expected reward/cost measure meet some threshold or is maximized (minimized).
	
		
Consider the PCTL property~(\ref{eqn:time}) that captures the high-level goal of our mobile robotics scenario.
\begin{equation}
\underbrace{\sf{R}\{{ \sf time}\}_{ \sf min=?}}_{\text{Reward quantifier}} \; \underbrace{ \sf{[ F \; { \sf goal} ]} }_{\text{Path formula}}
\label{eqn:time}
\end{equation}

This property contains two parts: (i)~a {\em reward/cost quantifier}, which indicates that the synthesis strategy should minimize the accrued reward {\small \sf time} as defined in Listing~\ref{lst:timereward}, and (ii)~a {\em path formula}, which indicates that the paths in the model over which the reward has to be optimized are those that lead to states where the reachability predicate {\small \sf goal} is satisfied. This predicate requires that the robot's final location is the target location for the mission.



\subsection{Integrated Learning and Adaptation}
\label{sec:integration}

\begin{figure}[t]
	\begin{center}
		\includegraphics[width=0.8\columnwidth]{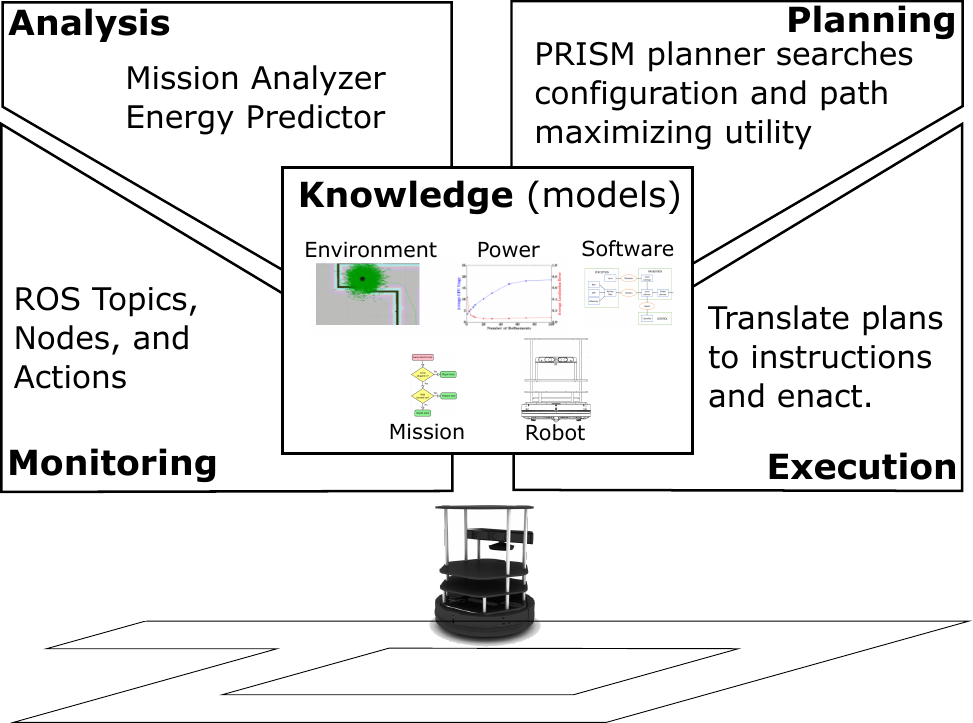}
		\caption{Using the learned model as the Knowledge in MAPE-K.}
		\label{fig:mape}
	\end{center}
\end{figure}

To use the model and planning synthesis previously described, we need to embed them in a control loop~\cite{CSSASS:TAAS17} that runs with the robot and monitors various aspects of the robot and the environment it encounters, and adapting the robot configuration to increase the likelihood that a mission will be successful.
To implement the approach described above, we apply Rainbow~\cite{Garlan2004,2012:Cheng:Stitch}, which is an extensible framework for developing self-adaptive systems, while broadly applies the self-adaptive elements of Modeling, Analysis, Planning, and Execution, with a Knowledge Base of models, referred to as MAPE-K~\cite{Kephart:2003:VAC}, and customized as shown in \figref{mape}:
\begin{itemize}
\item \textbf{Modeling and Analysis.} In addition to containing a model of the robotic software architecture, we also model the environment (locations of interest in a map, position of charging stations), mission (instructions for the robot, progress through the mission), robot state (location in the environment, speed, battery level), and power consumption models. The learned power model is consumed by Rainbow on start-up so that it can be used by the planner. Probes are inserted into the robotic software to get information into the model. 
ROS Kinetic, the robotics software that we use, has tools for monitoring the messages that are sent between various components of the robot (including sensed information), which are sufficient to update the models. The probes report both periodically (\emph{e.g.}, for location or charge) once every second, or when some event occurs. 

To interpret the model information and determine erroneous conditions requiring adaptation, we include two Analysis components. The first analysis is a \textit{Mission Analyzer} that determines if the mission is still on track. If the robot encounters an obstacle or the robot fails to execute some of its instructions, then this analyzer signals an error in the mission. The second analysis is an \textit{Energy Predictor} that uses the learned power model to determine if the robot has run out of power or if the robot does not have enough power to get to the charging station in the current plan (if there is one). If using the current state of the mission and the current charge in the battery there is not enough power to do this, then a new plan will need to be synthesized. 

\item \textbf{Planning.} Rainbow uses a utility-based strategy selection method to select from a fixed set of adaptations. In this work, we integrated the plan synthesizer discussed above that uses quantitative analysis to combine information from the multiple models and synthesizes plans that choose optimal mission improvements and reconfigurations to achieve mission goals as the environment changes (\emph{e.g.}, power consumption is significantly different to that which is predicted). When a problem is identified by one of the analyzers, the planner then synthesizes a plan which (a) indicates the robot configuration to use and (b) specifies the path that the robot should take in the form of a list of ordered waypoints.

\item \textbf{Execution.} We replace the Stitch strategy execution machinery that has previously been used (\emph{e.g.}, ~\cite{2012:Cheng:Stitch}) with tactics to issue new tasks to the robot and that reconfigure the robot. The plan generated by the planner is translated into an instruction graph~\cite{Mericli2014}, which is a sequence of instructions that the robot understands and is then forwarded to the robot to be executed. The current set of instructions is canceled and the new instruction graph is executed on the robot. 
\end{itemize}

\section{Implementation and Integration}
\label{sec:implementation}

To evaluate our approach, we implemented the testing infrastructure in Gazebo,\footnote{http://gazebosim.org/} which is used to create applications for a physical robot without depending on the actual machine. Our testing infrastructure is compatible with ROS and can be transferred onto the physical robot without modifications.

To enable evaluation, we have specified a rich API for an evaluator to exercise and explore the robustness of the system. 
We have implemented the following components and tools to evaluate our approach:

\begin{itemize}
	\item A \emph{test harness} that specifies the test configurations (start, target locations, number of tasks in a mission) and perturbations to the environment. The test harness was provided by an independent evaluator, MIT Lincoln Laboratory.
	\item A \emph{learning component} implementing the learning approach described in \secref{learning}. Our code for learning and extraction, including some tutorials, are available at \url{https://github.com/cmu-mars/model-learner/tree/tutorial}.
	\item A set of perturbation scripts to make run-time modification, \emph{e.g.}, placing obstacles, suddenly changing battery levels.
	\item A set of \emph{backend components} including robot models describing all aspects of Turtlebot, static and dynamic objects, lighting, terrain, and even physics of the robot. Also, several Gazebo plug-ins to implement configuration management of robots and power simulation. Source code can be found here: \url{https://tinyurl.com/ycketxbb}.
	\item An \emph{instruction graph}~\cite{Mericli2014} that allows the creation of task plans. These task plans can be transferred across different ROS-enabled robotics platforms. This component provides a standardized interface for interfacing with preemptable tasks, \emph{e.g.}, moving the robot to a target location.
\end{itemize}

\section{Evaluation}
\label{sec:evaluation}

To evaluate the two hypotheses formulated in \secref{introduction}, we designed and ran experiments using (i) controlled experiments in a theoretical setting with a large configuration space that were validated on the real system (to increase internal validity, explore scalability, and evaluate different environmental characteristics) as well as (ii) independent evaluation on a robotics scenario that was described in \secref{example} (to ensure external validity~\cite{SSA:ICSE15}). 

\subsection{Hypothesis 1 -- Learning identifies optimal configurations}

To evaluate whether the learning mechanism we described in \secref{learning} will find Pareto-optimal configurations, we ran an experiment in which we learned models by taking a small set of random samples (100 samples) for all 100 synthetically generated power models and measured (a) rank correlations between the learned model and the actual model as well as (b)~prediction accuracy of the learned model. Following the methodology in our previous work in \cite{JVKS:FSE18}, we categorized the 100 synthetic models into 3 categories of \emph{Easy, Medium}, and \emph{Hard} models based on the number of interaction terms that appears in the model. We also compared the results with CART (Classification and Regression Trees) as a popular learner in the literature~\cite{GCASW:ASE13}. We used Spearman's rank correlation coefficient as a measure to assess how well the learned model is likely to detect Pareto-optimal configurations. Spearman's correlation assesses monotonic relationships (whether linear or not). A perfect Spearman correlation occurs when the configurations ranked by the actual model are a perfect monotone function of the configurations ranked by the learned model. 

\begin{figure}[t]
	\begin{center}
		\includegraphics[width=\columnwidth]{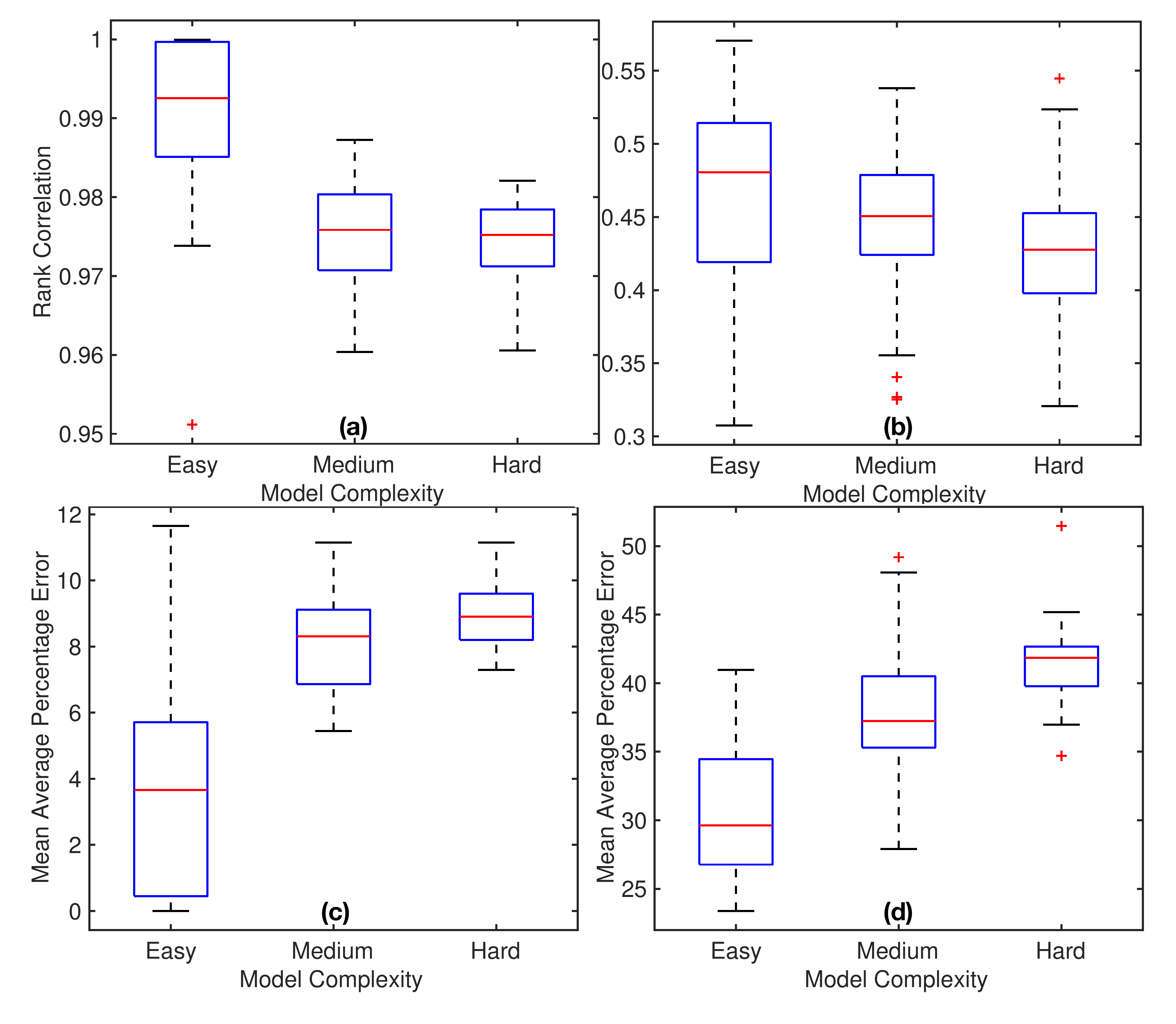}
		\caption{Rank correlation and accuracy of the model learning (a and c) and comparison with CART model (b and d). The results indicate that our approach is very likely to find Pareto-optimal configurations by only measuring performance of small portion of the configurations.}
		\label{fig:rank-accuracy-learning}
	\end{center}
\end{figure}

The results in \figref{rank-accuracy-learning} show that even with 100 samples (0.0001\% of the configuration space, it could be still a significant number depending on the size of the space, so we varied the number of samples~\cite{supplementary}), we are able to learn an accurate model (with prediction error of less than 10\% even for hard models) that is highly-likely (over 99\% for easy models and over 97\% for medium and hard models) to identify Pareto-optimal configurations. By comparing with a popular approach (CART) for model prediction in highly-configurable systems, the results indicate that our approach is able to find Pareto-optimal configurations even when only exploring a small portion of the configuration space while CART models were inaccurate (above 40\% error for Hard models) and were not able to find optimal configurations even for Easy models. The results for a sensitivity analysis can be found in supplementary material in~\cite{supplementary}.

\subsection{Hypothesis 2 -- Learning leads to higher quality}

We evaluate Hypothesis 2 using both controlled experiments and independent evaluation on a real-world system.

\subsubsection{Independent Evaluations (Settings)}
\label{sec:real}

For independent evaluation of our approach, MIT Lincoln Laboratory used the testing infrastructure that was described in \secref{implementation} to evaluate whether using our approach leads to higher quality missions of the robot over the use of no or static models. We characterize severe changes that might be caused by unanticipated and yet-unknown future environment changes in order to test the degree to which the system is able to support a higher-level navigation mission scenario by learning and adapting to these changes.


We used the scenario that was described in \secref{example}. 
In each test, the robot was asked to complete a mission consisting of $N$ tasks within a map. The learned power model is used by the planner, but the actual battery levels in the robot use the ground truth power model.  
The testing infrastructure supports two different environmental perturbations: (i) battery perturbations and (ii) obstacle perturbations. For each test, a list of tasks is generated by randomly selecting (i) the number of tasks and (ii) their associated locations in the map. Given the mission, the following parameters were randomly chosen: 
\begin{enumerate}
\item the power model of the robot---from a list of 100 synthetically generated power models with different complexity (\emph{i.e.}, each with different number of option interactions);
\item the ‘learning budget’, indicating the maximum queries allowed against actual power model during learning; and
\item the type of dynamic environmental perturbation. 
\end{enumerate}
For each parameter, we provided many different values to allow the evaluator to test how sensitive our technology is with respect to model complexity, learning budget, and environmental perturbations. 
The tests were run in three stages: 
\begin{itemize}
	\item \textbf{Baseline A (no perturbations)} represents the unadaptive target system operating in an unperturbed environment. 
	\item \textbf{Baseline B (perturbations and reactive planning)} represents the reactive target system operating in a perturbed ecosystem. This stage is used to confirm that the perturbation to the ecosystem actually threatens the target system's ability to achieve its intent (\emph{i.e.}, for the robot to reach its goal). In Baselines A and B, we used threshold-based reactive planning to go to the nearest charging station. 
	\item \textbf{Challenge (perturbations and quantitative planning)} represents an adaptive target system where the planner has knowledge of the mission, environment, and learned power model, operating in a perturbed ecosystem. This stage is required to maintain or recover the intended functionality. 
\end{itemize}

We provided a Docker containerized system with a REST interface for interacting with the test harness. MIT Lincoln Lab chose a set of experiments, with each experiment forming a triple containing a run in Baselines A, B, and the Challenge case. Lincoln Lab were able to run over 280 test triples from which only 120 tests were valid.
A test triple was deemed \emph{Invalid} if the following were satisfied:
\begin{itemize}
	\item Baseline A did not finish the mission or the perturbations in Baseline B were not severe enough to prevent it from completing the mission; or
	\item some condition occurred during test execution that precluded continuation of the test (\emph{i.e.}, occurrence of an \emph{Error}\footnote{Note that since the testing infrastructure comprises of an ecosystem of independently developed systems deployed over cloud and runs on ROS the chance of infrastructure errors is not zero.}) or precluded interpretation of the test results (\emph{e.g.}, due to missing or malformed data or logs). 
\end{itemize}


Each test was evaluated to determine whether the Challenge case (adaptation with the learned model) was successful, with the following verdicts:
\begin{itemize}
	\item \emph{Pass}: The Challenge case was able to adapt the robot to accomplish all tasks in the mission.
	\item \emph{Degraded}: The robot accomplished \emph{more} tasks in Challenge than in Baseline B (but not all of them).
	\item \emph{Fail}: The robot accomplished \emph{fewer} tasks in Challenge than in Baseline B.
	\item \emph{Inconclusive}: The test completed, but the test information was too uncertain to be used as the basis for an outcome. 
\end{itemize}


\subsubsection{Independent Evaluation (Results)}

\begin{figure}[t]
	\begin{center}
		\includegraphics[width=\columnwidth]{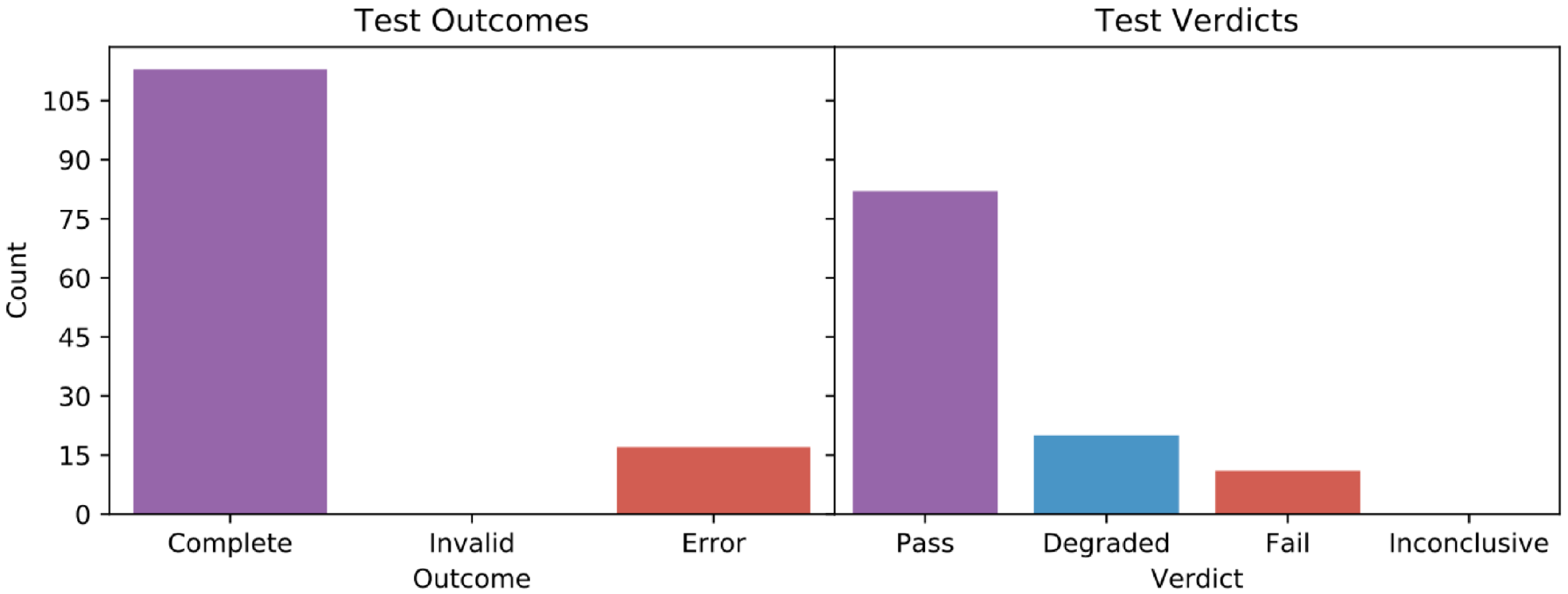}
		\caption{Results with obstacles perturbations. The number of Passes is significantly higher than Failures.}
		\label{fig:obstacle-perturb}
	\end{center}
\end{figure}

The results for obstacle test cases in \figref{obstacle-perturb} show that a high percentage of the tests produced valid outcomes (\emph{Complete}). Lincoln Lab were able to achieve this because of the way they constructed the tests. In selecting adjacent way points, they were careful to choose in such a way that when obstacles were placed between way points, they were highly confident that the robot would choose the path with the obstruction. Obstacles were also placed in such a way that it was possible to construct plans where the robot could achieve a task using an alternative path. The results show that the adapted system is able to recover functionality, only producing \emph{Degraded} and \emph{Fail} in about 18\% and 9\% of the test cases respectively.

\begin{figure}[t]
	\begin{center}
		\includegraphics[width=\columnwidth]{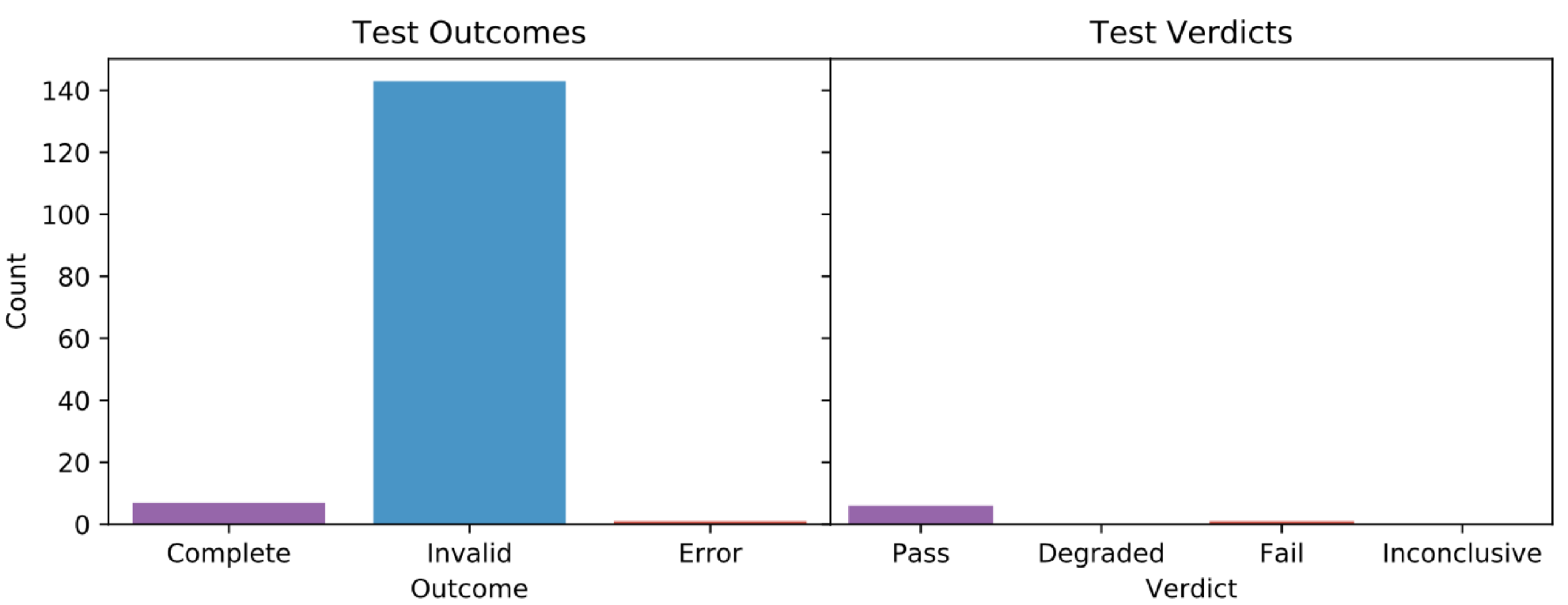}
		\caption{Results with battery perturbations. We faced many invalid test, most of which were Passed.}
		\label{fig:battery-perturb}
	\end{center}
\end{figure}

Generating valid test cases for the battery tests turned out to be a bit more challenging. The results in \figref{battery-perturb} show that a high number of tests were deemed \emph{Invalid}. The only way a test can fail is if the robot runs out of battery. The only way a robot can run out of battery is if it gets drained during traversal of the path or if we perturb the battery and drain it more quickly. Adding too many battery drain perturbation events will result in a test case where it is impossible for any system to recover (the robot would shutdown immediately). In contrast, having too little battery drain results in a test that is too `easy', so the Baseline B succeeds and it results in invalid test (based on our definition of a valid test). For the small number of tests that are valid, the result is promising, \emph{i.e.}, mostly passes with only a couple of tests failed.

\begin{figure}[t]
	\begin{center}
		\includegraphics[width=0.9\columnwidth]{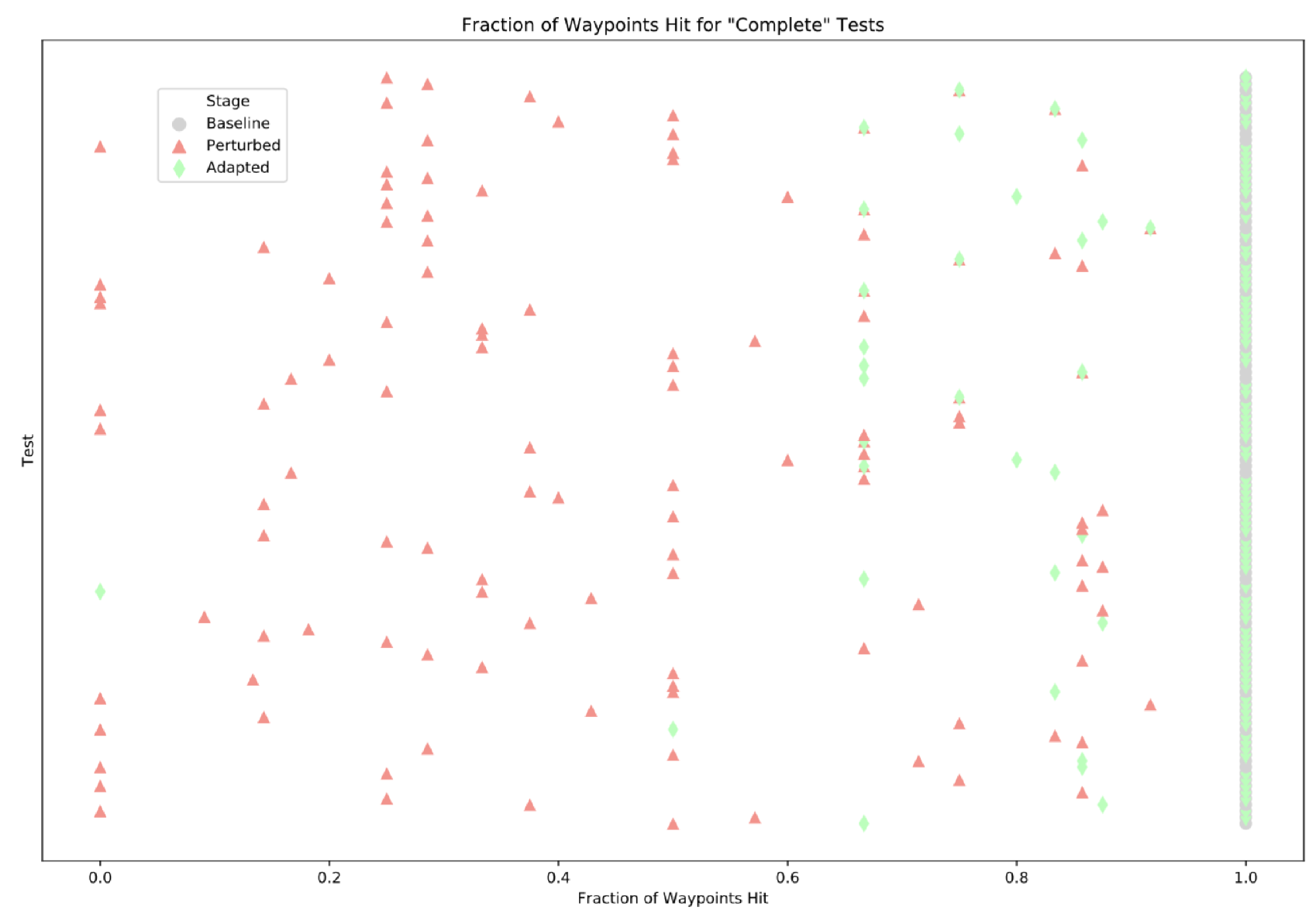}
		\caption{Task recovery as a function of test stage: our approach recovered more tasks comparing with baseline.}
		\label{fig:interpretation}
	\end{center}
\end{figure}

To look at the data from another angle, the results in \figref{interpretation} show the test identifier against the task completion ‘score’ metric. The score metric roughly represents the fraction of the tasks in the mission that the robot hit during the test execution. We plot this metric for completed tests. For the Baseline A stage, the robot successfully achieves all tasks (a score of 1.0). For the Challenge (Adapted) stage, the scores tend to be higher comparing with Baseline B, confirming that our approach can recover the mission.




\subsubsection{Controlled Experiment}

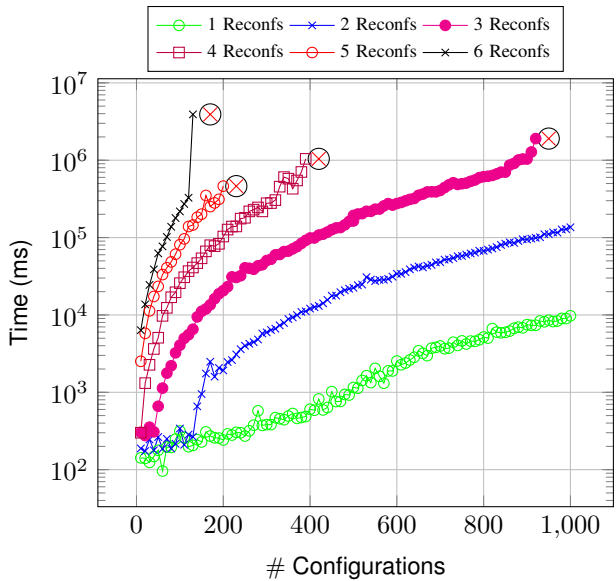
\begin{figure}[!h]
\begin{tikzpicture}
	\begin{axis}[
		legend style={font=\scriptsize},
		restrict y to domain=0:10000000,
		restrict x to domain=0:1000,
		legend columns=3,
		grid=major,
		xtick={0,200,...,1000},
		xlabel={$\#$ Configurations},
		ylabel={Time (ms)},
		legend style={at={(0.5,1.17)},anchor=north},
		ymode=log,
	]

\addplot[color=green,mark=o] coordinates {
(10, 142)
(20, 140)
(30, 124)
(40, 148)
(50, 178)
(60, 96)
(70, 196)
(80, 194)
(90, 248)
(100, 316)
(110, 260)
(120, 198)
(130, 206)
(140, 244)
(150, 228)
(160, 308)
(170, 272)
(180, 258)
(190, 256)
(200, 242)
(210, 290)
(220, 280)
(230, 304)
(240, 300)
(250, 272)
(260, 318)
(270, 376)
(280, 576)
(290, 374)
(300, 382)
(310, 382)
(320, 470)
(330, 456)
(340, 442)
(350, 494)
(360, 532)
(370, 460)
(380, 472)
(390, 486)
(400, 604)
(410, 586)
(420, 820)
(430, 592)
(440, 636)
(450, 1020)
(460, 760)
(470, 762)
(480, 938)
(490, 922)
(500, 1156)
(510, 1130)
(520, 1422)
(530, 1640)
(540, 1334)
(550, 2040)
(560, 1610)
(570, 1310)
(580, 1892)
(590, 1888)
(600, 2536)
(610, 2272)
(620, 2392)
(630, 2618)
(640, 2838)
(650, 3440)
(660, 3190)
(670, 2976)
(680, 3718)
(690, 3808)
(700, 3968)
(710, 3660)
(720, 3720)
(730, 4440)
(740, 4038)
(750, 4620)
(760, 4214)
(770, 4596)
(780, 4554)
(790, 4780)
(800, 5180)
(810, 5020)
(820, 6662)
(830, 6000)
(840, 5878)
(850, 5960)
(860, 6182)
(870, 6718)
(880, 6948)
(890, 6810)
(900, 7490)
(910, 7332)
(920, 7356)
(930, 8360)
(940, 8150)
(950, 8556)
(960, 8214)
(970, 8348)
(980, 9020)
(990, 8976)
(1000, 9746)
};
\addlegendentry{1 Reconfs}
\addplot[color=blue, mark=x] coordinates {
(10, 188)
(20, 174)
(30, 246)
(40, 178)
(50, 264)
(60, 186)
(70, 250)
(80, 188)
(90, 216)
(100, 340)
(110, 210)
(120, 284)
(130, 266)
(140, 658)
(150, 950)
(160, 1750)
(170, 2482)
(180, 1594)
(190, 2066)
(200, 1916)
(210, 2462)
(220, 2608)
(230, 3086)
(240, 3608)
(250, 4004)
(260, 4246)
(270, 4452)
(280, 4820)
(290, 5756)
(300, 5950)
(310, 6382)
(320, 6646)
(330, 7336)
(340, 7864)
(350, 8856)
(360, 9284)
(370, 9950)
(380, 10934)
(390, 11138)
(400, 12074)
(410, 12612)
(420, 13090)
(430, 14150)
(440, 15644)
(450, 17720)
(460, 17794)
(470, 19672)
(480, 20840)
(490, 21838)
(500, 22466)
(510, 24190)
(520, 24706)
(530, 30776)
(540, 28604)
(550, 27456)
(560, 28350)
(570, 28496)
(580, 29274)
(590, 31246)
(600, 34012)
(610, 34028)
(620, 35798)
(630, 38800)
(640, 40910)
(650, 42314)
(660, 41180)
(670, 43326)
(680, 44432)
(690, 47770)
(700, 48058)
(710, 51858)
(720, 52262)
(730, 54430)
(740, 57426)
(750, 57916)
(760, 60162)
(770, 62342)
(780, 63930)
(790, 67536)
(800, 67276)
(810, 69934)
(820, 72254)
(830, 75280)
(840, 80288)
(850, 83494)
(860, 86770)
(870, 85476)
(880, 90294)
(890, 95124)
(900, 94532)
(910, 97024)
(920, 99350)
(930, 101000)
(940, 110520)
(950, 110926)
(960, 116760)
(970, 116050)
(980, 127156)
(990, 128746)
(1000, 135878)
};
\addlegendentry{2 Reconfs}

\addplot[color=magenta,mark=*] coordinates {
(10, 304)
(20, 276)
(30, 354)
(40, 308)
(50, 660)
(60, 1132)
(70, 1770)
(80, 2204)
(90, 3238)
(100, 4064)
(110, 4984)
(120, 5556)
(130, 6540)
(140, 9436)
(150, 11124)
(160, 11956)
(170, 13534)
(180, 16206)
(190, 18774)
(200, 20618)
(210, 23178)
(220, 31042)
(230, 30482)
(240, 32128)
(250, 40664)
(260, 40092)
(270, 38958)
(280, 43138)
(290, 44542)
(300, 50446)
(310, 52484)
(320, 59784)
(330, 60484)
(340, 65936)
(350, 66862)
(360, 71504)
(370, 76520)
(380, 84060)
(390, 91984)
(400, 98714)
(410, 98928)
(420, 107720)
(430, 110724)
(440, 117730)
(450, 126544)
(460, 133296)
(470, 134714)
(480, 146072)
(490, 161796)
(500, 163974)
(500, 193268)
(510, 201448)
(520, 203858)
(530, 218818)
(540, 218102)
(550, 232770)
(560, 233986)
(570, 256632)
(580, 274722)
(590, 263092)
(600, 275618)
(610, 284716)
(620, 298584)
(630, 311774)
(640, 320304)
(650, 352336)
(660, 365436)
(670, 388852)
(680, 390516)
(690, 389980)
(700, 406276)
(710, 443252)
(720, 480594)
(730, 511652)
(740, 486498)
(750, 496796)
(760, 506804)
(770, 533800)
(780, 555474)
(790, 595234)
(800, 613222)
(810, 620808)
(820, 635578)
(830, 664694)
(840, 701662)
(850, 701458)
(860, 864870)
(870, 909560)
(880, 1006698)
(890, 1036596)
(900, 1035768)
(910, 1276896)
(920, 1904610)
};
\addlegendentry{3 Reconfs}
\addplot[color=purple,mark=square] coordinates {
(10, 300)
(20, 1314)
(30, 2266)
(40, 3632)
(50, 5076)
(60, 9728)
(70, 12354)
(80, 16934)
(90, 19632)
(100, 25740)
(110, 30716)
(120, 36838)
(130, 41022)
(140, 46790)
(150, 53974)
(160, 66388)
(170, 79716)
(180, 77492)
(190, 83088)
(200, 103258)
(210, 126938)
(220, 138920)
(230, 139432)
(240, 175722)
(250, 180220)
(260, 218092)
(270, 229642)
(280, 245132)
(290, 218042)
(300, 277930)
(310, 279506)
(320, 305234)
(330, 454980)
(340, 605244)
(350, 577295)
(360, 428172)
(370, 547378)
(380, 704716)
(390, 1041574)
};
\addlegendentry{4 Reconfs}
\addplot[color=red,mark=o] coordinates {
(10, 2516)
(20, 5780)
(30, 11296)
(40, 17326)
(50, 23378)
(60, 33242)
(70, 40706)
(80, 48528)
(90, 60986)
(100, 80890)
(110, 96222)
(120, 139034)
(130, 145370)
(140, 183272)
(150, 200864)
(160, 350130)
(170, 251592)
(180, 280614)
(190, 312222)
(200, 462520)
};
\addlegendentry{5 Reconfs}
\addplot[color=black,mark=x] coordinates {
(10, 6350)
(20, 13682)
(30, 24430)
(40, 38788)
(50, 62524)
(60, 76916)
(70, 101904)
(80, 137874)
(90, 179012)
(100, 217950)
(110, 272190)
(120, 327302)
(130, 3910212)
};
\addlegendentry{6 Reconfs}

\draw (950, 1904610) circle (4pt);
\draw (950, 1904610) node[cross,red] {};	
\draw (420, 1041574) circle (4pt);
\draw (420, 1041574) node[cross,red] {};	
\draw (230, 462520) circle (4pt);
\draw (230, 462520) node[cross,red] {};	
\draw (170,3910212) circle (4pt);
\draw (170,3910212) node[cross,red] {};	
\end{axis}

\end{tikzpicture}
\caption{Planning times for different configuration set sizes and number of allowed reconfigurations.}
\label{fig:planningtimes}
\end{figure}

To assess whether restricting planning to a Pareto-optimal set of configurations makes planning tractable at run time, we ran an experiment in which we supply sets of configurations of increasing size to our planner (Figure~\ref{fig:planningtimes}) and observe (a)~planning time and (b)~whether the planner was able to come up with a solution (or run out of memory instead). The planner's backend is based on Prism 4.3's explicit state space engine running on 8GB of allocated memory, macOS 10.14.1, and a 2.8GHz Intel Core i7. In the figure, the horizontal axis displays the number of configurations, and the vertical axis displays time in milliseconds (on a logarithmic scale) that it takes to compute a plan for the path going from {\sf l1} to {\sf l5} in the map displayed in Figure~\ref{fig:scenario}. We plot times for an increasing number of reconfigurations allowed, ranging from one to six. 

The results show that very moderate numbers of configurations considered ($10^3$ is only a tiny fraction of the overall configuration space) already lead to long planning times (about 10 seconds even when only one reconfiguration is allowed). 
This is aggravated in the case of two reconfigurations, in which planning time jumps to over 100 seconds, and higher numbers of allowed reconfigurations exhibit a similar exponential increase in computation time as well as the exhaustion of available memory (marked by red crosses in the figure). 
However, if we keep the number of configurations supplied to the planner in the range 10--180, we can observe how computation times are kept below 10 seconds, even when three reconfigurations are allowed. 
This indicates that adjusting Pareto-optimal set sizes can lead to tractable run-time planning under a different number of allowed reconfigurations, achieving tradeoffs that depend on the specific timeliness requirements of the adaptation scenario. 


\subsection{Threats to Validity}

\subsubsection*{Internal and Construct Validity}
To ensure internal validity of our approach, the evaluations were conducted independently by a third party evaluator. We defined and implemented an API for interacting with the experimental platform and Lincoln Laboratory used the API to run the tests in their internal testing infrastructure. The results of evaluation were directly reported to DARPA, the project funding agency, without any intervention from the researchers.  
The robustness guarantees provided by the quantitative planning is partially dependent on the accuracy of the data provided by machine learning.

\subsubsection*{External Validity}
Even though we expect that the integrated model learning and quantitative planning would be beneficial for any configurable systems, the results need to be interpreted only in the context of robotics systems. For increasing external validity, we used 100 synthetically generated power models. 

\section{Related Work}
\label{sec:related-work}

\textbf{Configuration optimization}.
Configurations can affect functional and extra-functional aspects of systems. The performance of these systems depends on configurations~\cite{SGAK:ESECFSE15}.
Several models (\emph{e.g.}, support-vector~\cite{YWLE:MASCOTS13}, decision trees~\cite{NMSA:Arxive17}, Fourier sparse functions~\cite{ZGBC:ASE15}), sampling strategies (\emph{e.g.}, active learning~\cite{SGAK:ESECFSE15}), and optimization (\emph{e.g.}, search-based and evolutionary algorithms~\cite{HPHL:ICSE15,WWHJK:GECCO15}) have been used for performance optimization of highly-configurable systems~\cite{WFP:FOSE07}. 
Configuration optimization search for optimal configurations given a limited budget considered recursive random sampling~\cite{YK:SIGMETRICS}, hill climbing~\cite{XLRXZ:WWW04}, direct search~\cite{ZBN:OSR}, optimization via guessing~\cite{OK:SIGMETRICS07}, Bayesian optimization~\cite{JC:MASCOTS16}, and multi-objective optimization~\cite{FHM:FSE15}, and ablation analysis~\cite{BLEHFH:AAAI17}.

In this work, we learn performance models~\cite{JVKS:FSE18,SGAK:ESECFSE15}. We then use the learned model to identify Pareto-optimal configurations to enable quantitative planning for self-adaptation of highly-configurable autonomous robots. However, as opposed to our previous work~\cite{JVKSK:SEAMS17}, we do not perform any particular experiments with transfer learning involved. 

\textbf{Planning.}
There exist approaches on finding solutions to adaptations based on planning and constraint solving~\cite{zeller2012timing}. Also, statistical techniques have been used for providing guarantees in self-adaptive systems~\cite{weyns2016model} with focus on real-time constraints~\cite{rodrigues2018learning} and continuous learning~\cite{buttar2018applying}. 

Existing approaches that enable self-adaptation of systems rely on quantitative verification of system properties, including techniques that can be used to produce formal guarantees about quantitative aspects of systems, such as performance~\cite{DBLP:conf/sigsoft/MorenoCGS15,DBLP:conf/aips/Lacerda0H17,DBLP:conf/sac/CamaraGS015} as well as appropriate reference frameworks~\cite{braberman2015morph}.
For example, probabilistic model checking is typically used to evaluate the result of
queries based on an analysis of a probabilistic model of the system behavior. This requires exhaustive explorations to check for all possible executions and then queries are solved
via numerical methods~\cite{DBLP:conf/sigsoft/MorenoCGS15,DBLP:conf/aips/Lacerda0H17}.

These approaches perform exhaustive search to verify all the adaptation options to find the best decision. These analyses happen at run time and the models are often simplified because this is a very computationally demanding task. The time and resource demands are directly dependent on the size of the adaptation space. Therefore, such approaches cannot be used in resource-constrained devices. However, our approach restricts the quantitative planning to the Pareto-optimal configurations, therefore, enabling run-time self-adaptations for highly-configurable autonomous robots.

\textbf{Machine learning for run-time decision making.}
There are studies that use machine learning to support decision making at run time. 
Performance reasoning is a key activity. Time series techniques~\cite{ehlers2011self} have proven to be effective in predicting response time and uncovering performance anomalies. FUSION~\cite{esfahani2013learning,elkhodary2010fusion} exploited inter-feature relationships to reduce the dimensions of configuration spaces making run-time performance reasoning feasible. Different classification models have been evaluated for the purpose of time series predictions~\cite{anaya2014prediction}.
Performance predictions have also been applied for resource allocations~\cite{huber2017model,JPM:Cloud} or other related applications~\cite{filieri2015automated,becker2006performance}.
Note that the approaches above are referred to as black-box models. However, another category of models known as white-box can be built early in the life cycle, by studying the underlying architecture of the system \cite{gomaa2007model,happe2011facilitating} using Queuing networks and Petri Nets~\cite{balsamo2004model}. Constraining the configuration space has also been explored in other types of systems, \emph{e.g.}, service-oriented applications~\cite{epifani2009model,calinescu2011dynamic,morin2009models}.

These approaches, especially FUSION~\cite{esfahani2013learning,elkhodary2010fusion}, bear resemblance to our work in their use of learning. While they have been shown to be useful, they suffer from scalability (the adaptation space did not exceed to more than a thousand alternatives, while our approach is tested for over one million configurations). 


\section{Conclusions}
\label{sec:conclusions}

We proposed an integrated learning and quantitative planning approach that enables adaptation of robots. The key novelty of our approach is the integration of learning and quantitative planning to enable run time self-adaptations, and integration of information from multiple heterogeneous models in quantitative planning. Unlike most other research in adaptive systems community, our work was evaluated by an independent party. As future work, we envisage online model learning to account for model update at run time. 



\section*{Acknowledgments}
This material is based on research sponsored by \textbf{AFRL} and \textbf{DARPA} under agreement number FA8750-16-2-0042. This work has been also supported in part by the \textbf{National Science Foundation} (awards 1318808, 1552944, and 1717022). The authors would like to thank the \textbf{CMU BRASS MARS} team members especially Miguel Velez (for analysis component development), Ivan Ruchkin (for power model inspiration and checking), Roykrong Sukkerd, Ashutosh Pandey (for partial Rainbow development), Ian Voysey (for integration with Lincoln Labs test environment), Timothy Braje,  Jeffrey Hughes, Timothy Meunier, W. Konrad Vesey (for evaluation and help with problem definition for the evaluation), Jonathan Aldrich (for being the lead PI). We also thank Marilyn Gartley for proofreading the final draft.


\bibliographystyle{abbrv}
\bibliography{bibliography}

\end{document}